# Retrieval Augmented Generation using Engineering Design Knowledge

L. Siddharth[1✉], Jianxi Luo[2]

[1]*Engineering Product Development, Singapore University of Technology and Design, Singapore*

[2]*Department of Systems Engineering, City University of Hong Kong, Hong Kong*

## Abstract

Aiming to support Retrieval Augmented Generation (RAG) in the design process, we present a method to identify explicit, engineering design facts – {*head entity :: relationship :: tail entity*} from patented artefact descriptions. Given a sentence with a pair of entities (based on noun phrases) marked in a unique manner, our method extracts the relationship that is explicitly communicated in the sentence. For this task, we create a dataset of 375,084 examples and fine-tune language models for relation identification (token classification) and elicitation (sequence-to-sequence). The token classification approach achieves up to 99.7% accuracy. Upon applying the method to a domain of 4,870 fan system patents, we populate a knowledge base of over 2.93 million facts. Using this knowledge base, we demonstrate how Large Language Models (LLMs) are guided by explicit facts to synthesise knowledge and generate technical and cohesive responses when sought out for knowledge retrieval tasks in the design process.

## 1. Introduction

**1.1. Motivation.** Large-Language Models (LLMs) such as GPT, Llama, Mistral, Falcon etc., are increasingly being adopted for applications like chatbots, writing assistants, data curators etc. Despite growing popularity, LLMs are less suited for knowledge-intensive tasks in engineering design, e.g., case-based reasoning (Qin et al., 2018; Siddharth et al., 2020). These tasks are typically supported by designated or hand-built knowledge sources like handbooks, patents, technical reports etc., (Quintana-Amate et al., 2015; Siddharth et al., 2022b). For LLMs to generate responses within the context of such sources, it is necessary to devise a method to extract engineering design knowledge from these. In this paper, we propose a method to populate facts from artefact descriptions found in patent documents.

**1.2. Methodology**. A fact is represented as a triple - *head entity :: relationship :: tail entity*. As illustrated in Figure 1, the engineering design facts in a sentence[1] could be identified using explicit relationships that associate a pair of entities (noun phrases). The entities are various constituents of an artefact, and the relationships are used to describe its underlying structures, behaviours, and purposes (Chandrasekaran, 2005; Siddharth et al., 2018). For a given pair of entities (as noun phrases), e.g., "processes" and "Progesterone Receptor" in Figure 1, the objective of our method is to return the explicit relationship - "mediated by". To this end, as explained in Section 3.1, we create a large dataset of 44,227 sentences and corresponding facts from a stratified sample of 4,205 patents granted by the USPTO[2].

✉ Corresponding Author. siddharth_l@mymail.sutd.edu.sg, siddharthl.iitrpr.sutd@gmail.com.

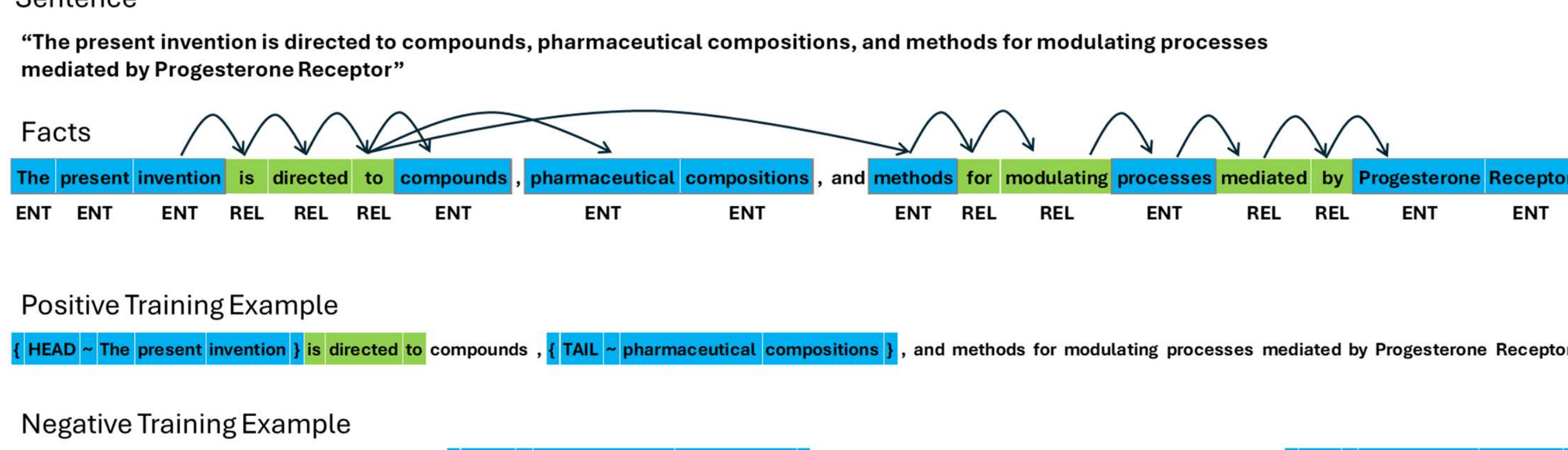

**Figure 1**: Illustrating our approach for relation extraction.

Sentences in the dataset may include one or more facts as illustrated in Figure 1. For each fact, e.g., *The present invention :: is directed to :: pharmaceutical compositions*, we create a positive training example as illustrated in Figure 1. When a sentence with entities marked as in Figure 1 is given as input, the proposed method is expected to return "is directed to". For a negative example as in Figure 1, where a pair of entities do not have an explicit relationship in the sentence, the proposed method is expected to return NIL. We process the dataset in this manner to populate 375,084 training examples.

To train models using the processed dataset, we undertake two broad approaches as detailed in Section 3.2.

1. *Relation identification* – For a sentence with entities marked as in Figure 1, we fine-tune language models for token classification, i.e., to predict token labels (or tags) such as "HEAD", "REL", "TAIL", and "OTH" among which "REL" tokens shall be retrieved as the relationship.
2. *Relation elicitation* - For a sentence with entities marked as in Figure 1, we also fine-tune language models for Seq2Seq or Text2Text generation task, i.e., to directly output the actual relationship between the pair of entities.

**1.3. Contributions**. Based on the research, as reported in this paper, we make the following contributions to engineering design and knowledge engineering literature.

- We offer a new perspective of knowledge extraction, specific to engineering design applications and radically different from conventional knowledge extraction algorithms (as reviewed in Section 2.1).
- Visually demonstrated on YouTube[3], we manually identify engineering design facts from over 50,000 sentences by adopting several heuristics tailored to patent documents that are standard artefact descriptions.
- Upon processing the facts thus identified, we provide an open-source dataset of 375,084 training examples (187,200 positive, 187,884 negative) that is accessible on Hugging Face[4].
- For the dataset as above, we also provide the training infrastructure on Zenodo[5] that includes Python notebooks and the dataset formatted according to token classification and Seq2Seq tasks.
- We provide access to the best-performing model on Hugging Face[6] and facilitate usage of the method using readable instructions and an example notebook on GitHub[7].
- Upon applying the method to a domain of fan system patents and populating an accessible knowledge base[8], as explained in Section 4, we demonstrate Retrieval-Augmented Generation[9] (RAG) using explicit domain facts.

## 2. Background

**2.1. Knowledge Extraction Algorithms.** Knowledge extraction is a family of algorithms that involve two major tasks with a piece of text. First, Named Entity Recognition (NER), where phrases (also denoted as span) are identified and classified into entities. Second, predicting a relationship between a pair of entities thus classified (Li et al., 2022). These tasks are trained on standard datasets like ACE[10], W-NUT[11], and OntoNotes[12]. ACE, for example, comprises tagged broadcast transcripts, and news data in multiple languages including English. The entity classes in ACE include Person, Organization, Location, Facility, Weapon, Vehicle and Geo-Political. The relationships include physical (e.g., near, part-whole), social (e.g., business, family), socio-physical (e.g., ownership), affiliations etc.

NLP scholars and practitioners have proposed several generic algorithms for entity and relation extraction. Conventional relation extraction algorithms stem from syntactic dependencies between a pair of tokens or phrases, also denoted as subject-object pairs. For example, Sun and Grishman (2022) learn relations between entities using the lexical dependency paths. Due to the efficacy of Language Models, recent algorithms adopt these for encoding entities and relationships. Zhong and Chen (2021) fine-tune BERT for classifying spans into entity categories. Given a pair of spans, they include start and end markers with entity category and fine-tune BERT for predicting relation classes.

As a subject can be related to many objects, Ye et al. (2022) mark all objects tied to a subject and fine-tune BERT, ALBERT, and SciBERT for relation prediction. Liu et al. (2021) capture relationships among entities that are placed in different sentences in a document. They gather BERT-based token embeddings of entities and couple these using CNNs to predict relationships. Geng et al. (2023) map a sentence to a 2D representation and train Bi-TDNN to learn dependencies between entities (span of tokens). While sophisticated algorithms are continuously being proposed, stable ones are also being made available in NLP toolkits such as spaCy[13], TextRazor[14] that enable direct usage of entity and relation extraction models.

**1.3. Engineering Design Knowledge**. Engineering design literature reports several NLP applications and ontologies (Siddharth et al., 2022a) for entity and relation extraction. Trappey et al. (2014) use TF-IDF values to extract keyphrases from patent documents and research articles concerning dental implants. They (2014, p. 158) build a domain ontology by linking these using relationships such as "type of", "consist of", "the same as", "has shape of" etc. Yang et al. (2018) compile 114,793 reports pertaining to the verification of digital pre-assembly of vehicle components to extract meaningful n-grams (e.g., brake pad) and associate these using relationships such as "is A" and "is Same To". To aid in biologically inspired design, Chen et al. (2021) obtain sentences from the AskNature database and tie syntactic dependencies (e.g., nsubj, dobj, nsubjpass) to constructs such as structure, function, and structure-function. Jang et al. (2021) use syntactic dependencies and BERT-based similarities to form a network of nouns and verbs B60 patents.

Dimassi et al. (2021) propose HERMES ontology for 4D printing that involves relationships such as "is A", "Is Composed Of", "has Spatial Relation", "has", "is Manufactured By" etc. Fatfouta and le Cardinal (2021) propose an ontology for car crash simulation using relationships such as "has", "solves", "belongs to" etc. Sun et al. (2022) populate a design knowledge graph from patents using rules applied to subject-verb-object triples obtained through dependency parsing. To aid fault-detection in additive manufacturing, by searching relevant articles, Wang and Cheung (2023) populate a domain knowledge graph of 127 facts constituted by the relationships "Influence Positive", "Influence Negative", "Is Composed Of", "Result In" etc., and feed into a graph convolutional network to discover new edges to inform defects.

Knowledge extraction in engineering design literature has largely been ontological, drawing from the ideologies of typical knowledge extraction in core NLP literature. Scholars in engineering design attempt to approximate a domain using a finite set of relationships. This approach might be meaningful in common-sense knowledge extraction, where relationships like "near", "family" are sufficient to overview a large body of information populated on the internet. The common-sense knowledge graphs that result from such approximations appear to serve well for keyword search and recommendation systems (Chang et al., 2023; Feng et al., 2021).

Engineering design text includes artefact descriptions and various aspects thereof. These descriptions mainly communicate the constituents of the artefacts and the structural and behavioural relationships among these (Chandrasekaran, 2005; Siddharth et al., 2018). Such relationships could range from simple ones like "comprise" to complex ones like "positioned to deviate from". The above-reviewed efforts tend to approximate these relationships into a finite set of alternatives as it is common in knowledge engineering literature. Our work departs from this notion and proposes a method to identify explicit facts in a sentence.

Engineering design scholars attempt to extract knowledge from documents like pre-assembly verification and simulation reports that are highly contextualised to an organisation or an artefact. Since our work is not focused on a particular domain, we develop the method to extract knowledge from patent documents that provide standard artefact descriptions for over 8 million artefacts granted by USPTO as of 2024. In our earlier work, we attempted to identify facts from patent claims using a simple rule-based approach (Siddharth et al., 2021). In this paper, we report an advanced, data-driven approach to populate facts from entire descriptions (not just claims).

## 3. Method

**3.1. Dataset**. Patents have several classifications (refer to CPC scheme[15]) and explain artefacts at different levels in detail in each section (refer to example patent[16]). To create a dataset for extracting facts, we sample sentences from USPTO considering these two variations. First, we gather a sample of 4,205 patents stratified according to 3-digit classes (e.g., F17 - Storing or Distributing Gases or Liquids) in the CPC scheme. Second, we scrape one paragraph from each document section from each patent in the sample. To gather the sample of patents, we acquire metadata of 7.9 million patents from PatentsView[17] (accessed on Jan 10, 2022). Upon filtering utility patents and integrating domain information, we obtain over 4.8 million patents from which we select an appropriate sample. We use Cochran's (1977) formula for sample size calculation as follows.

$$S = \frac{Z^2 \cdot p(1-p)}{E^2} \tag{1}$$

Where $S$ is the sample size, $Z$ is the z-score that varies according to the desired confidence level, $p$ is the proportion of the population, and $E$ is the margin of error. As the above formula returns the same sample size for all population sizes, it could be corrected for a specific population size as follows.

$$S' = \frac{S}{1 + \frac{S}{N}} \tag{2}$$

Where $S'$ is the corrected sample size according to the population size $N$. For the population size of 4,826,485, applying typical inputs such as 95% confidence and 5% error, the sample size turns out to be 385 that is quite small. Upon multiple trials, we observe a sample size of 4,141 for 99% confidence and 2% error. By proportionately sampling from each class with a minimum of one patent per class, the final sample includes 4,205 patents.

To gather one paragraph from each document section of each patent in the sample, we use the Beautiful Soup library[18] and scrape the web documents given in Google Patents[19]. From these paragraphs, we retain sentences within 100 words (approximately 200+ transformer-based tokens). The scraped text from the sample includes 51,232 sentences. To identify facts from the scraped text as illustrated in Table 1, we develop and utilise a web-based tagging interface[3] that allows interactive recording of facts. Prior to identifying facts, in the interface, we discard the patent document sections[20] that do not reflect artefact-specific knowledge, e.g., background, examples, drawing descriptions etc. We also discover undesirable lexical and syntactic characteristics, e.g., the occurrence of "." in "FIG. 2", "mg.", and "%." causing line breaks, and text within braces (.), [.], {.}, <.> that detection of noun-phrases and syntactic dependencies.

As identifying facts requires a deep understanding of entities and relationships from an engineering design perspective, this task could not be outsourced to laypersons and was primarily carried out by an engineering design scholar, harnessing intermittent guidance from more experienced scholars in design theory, data-driven design, and NLP techniques. We provide over 120 examples of sentences (with facts) from the dataset externally[21], in addition to the representative ones indicated in Table 1.

**Table 1**: Representative examples of sentences and facts. The entities are chosen from noun phrases (identified by spaCy Transformers) and **relationships** are selected from the remaining tokens given a pair of entities.

| **Hierarchy** | |
|---|---|
| *"The catalyst component **comprises** a magnesium chloride support, a titanium trichloride, an aluminum trichloride and an organic phosphite"*[22] | *The catalyst component :: comprises :: a magnesium chloride support*<br>*The catalyst component :: comprises :: a titanium trichloride*<br>*The catalyst component :: comprises :: an aluminum trichloride*<br>*The catalyst component :: comprises :: an organic phosphite* |
| *"…the method **comprises generating** magnetic nanoclusters from a target of magnetic material, **crystallizing** the magnetic nanoclusters, and **depositing** the magnetic nanoclusters…"*[23] | *the method :: comprises generating :: magnetic nanoclusters*<br>*the method :: comprises crystallizing :: the magnetic nanoclusters*<br>*the method :: comprises depositing :: the magnetic nanoclusters* |
| **Entity Attributes** | |
| *"The protector 9 **has** curved walls 28, 29, guide rail 30, slider 31…"*[24] | *The protector :: has :: curved walls*<br>*The protector :: has :: guide rail*<br>*The protector :: has :: slider* |
| *"The pressure range **of** the cluster forming chamber 410, the encapsulation chamber 429 and the deposition chamber 430…"* | *The pressure range :: of :: the cluster forming chamber*<br>*The pressure range :: of :: the encapsulation chamber*<br>*The pressure range :: of :: the deposition chamber* |
| **Behaviours** | |
| *"High vacuum, which **indicates** minimal gas consumption, increases the size…"*[25] | *High vacuum :: indicates :: minimal gas consumption*<br>*High vacuum :: increases :: the size* |
| *"Organic ring compounds, para aminophenylacetic acid and 4 aminoantipyrine may also be **added to** the bath together…"*[26] | *4 aminoantipyrine :: added to :: the bath*<br>*para aminophenylacetic acid :: added to :: the bath*<br>*Organic ring compounds :: added to :: the bath* |
| **State Transitions** | |
| *"…under a condition that changes the growth mode **from** the substantially 3D growth mode **to** a substantially 2D growth mode"*[27] | *the growth mode :: from :: the substantially 3D growth mode*<br>*the growth mode :: to :: a substantially 2D growth mode* |
| *"In a write/read mode, data **is transited from** an input/output buffer **via** the transition circuit **to** an input of the write amplifier…"*[28] | *data :: is transited from :: an input/output buffer*<br>*data :: is transited via :: the transition circuit*<br>*data :: is transited to :: an input* |
| **Intricate Relationships** | |
| *"…said molding plates bound a mold cavity, **into** which a material can be **injected by** means of the injection molding device **during** operation…"*[29] | *a material :: injected into :: a mold cavity*<br>*a material :: injected by :: means*<br>*a material :: injected during :: operation* |
| *"The actuator is divided into two cells by a power transfer shaft **to** which **is attached** a vent follower"*[30] | *a vent follower :: is attached to :: a power transfer shaft* |
| *"…subdivided into three parts, a lid with a hole **through** which the shaft of the tap **extends**, an upper half…"*[31] | *the shaft :: extends through :: a hole* |
| *"Method of treating or preventing the inflammatory response of colitis in a subject* ***comprising administering to*** *the subject an effective amount of a substance"*[32] | *Method :: comprising administering :: an effective amount*<br>*an effective amount :: to :: the subject* |
| **Exemplars** | |
| *"…the milling can take place in the presence of a dry hydrocarbon diluent* ***such as*** *hexane, heptane, cyclohexane…"*[33] | *a dry hydrocarbon diluent :: such as :: hexane*<br>*a dry hydrocarbon diluent :: such as :: heptane*<br>*a dry hydrocarbon diluent :: such as :: cyclohexane* |
| **Introduction** | |
| *"The present invention generally* ***relates to accessing*** *one or more file systems, for example* ***scanning*** *files in a computer or computer system, and more* ***particularly to*** *a method, system and computer program"*[34] | *The present invention :: relates to accessing :: one or more file systems*<br>*The present invention :: relates to scanning :: files*<br>*The present invention :: relates particularly to :: a method*<br>*The present invention :: relates particularly to :: system*<br>*The present invention :: relates particularly to :: computer program* |
| **Compound Entities** | |
| *"An alternative is to read a write power setup value ΔP for the optical disc…"*[35] | *a write power setup value :: :: ΔP* |

The hierarchical relationships are given by "include", "comprise" etc., and attribute relationships as in "has" or "of" are quite common. Verbs are often used in combination with prepositions to represent behavioural relationships. The relationships including "from" and "to" express a transition from one state to another. Some examples show the tail entity is placed before the head entity and the relationships are intricately placed in text. For example, although "administering" and "to" are placed alongside each other, these tokens belong to separate facts. The additional examples that follow showcase how entities are exemplified, introduced, and combined.

**3.2. Relation Extraction**. While populating the dataset of over 50,000 sentences, we selected pairs of entities among the noun phrases and relationships from the remaining tokens. From this dataset, we formulate training examples as explained using Figure 1, wherein, for each fact in a sentence, we mark a pair of entities using *{HEAD ~ …}* and *{TAIL ~ …}*. Existing knowledge extraction algorithms have adopted innovative marking strategies as well (Soares et al., 2019; Zhang et al., 2019; Zhong and Chen, 2021). Using the examples thus formulated, we fine-tune language models to predict the relation using token tags (relation identification) or as text output (relation elicitation). These two training approaches are depicted in Figure 2.

**A - Relation Identification**

Training Example

| The | receiver | section | amplifies | { | HEAD | ~ | an | inbound | RF | signal | } | to | produce | { | TAIL | ~ | an | amplified | inbound | RF | signal | } |
|---|---|---|---|---|---|---|---|---|---|---|---|---|---|---|---|---|---|---|---|---|---|---|
| OTH | OTH | OTH | OTH | HEAD | HEAD | HEAD | HEAD | HEAD | HEAD | HEAD | HEAD | REL | REL | TAIL | TAIL | TAIL | TAIL | TAIL | TAIL | TAIL | TAIL | TAIL |

Training Input

| [CLS] | the | receiver | section | amp | ##li | ##fies | { | head | ~ | an | in | ##bound | rf | signal | } | to | produce | { | tail | ~ | an | amplified | in | ##bound | rf | signal | } | [SEP] |
|---|---|---|---|---|---|---|---|---|---|---|---|---|---|---|---|---|---|---|---|---|---|---|---|---|---|---|---|---|
| 101 | 1996 | 8393 | 2930 | 23713 | 3669 | 14213 | 1063 | 2132 | 1066 | 2019 | 1999 | 15494 | 21792 | 4742 | 1065 | 2000 | 3965 | 1063 | 5725 | 1066 | 2019 | 26986 | 1999 | 15494 | 21792 | 4742 | 1065 | 102 |

Transformer Model

Predicted Output

| -100 | 3 | 3 | 3 | 3 | 3 | 3 | 0 | 0 | 0 | 0 | 0 | 0 | 0 | 0 | 0 | 1 | 1 | 2 | 2 | 2 | 2 | 2 | 2 | 2 | 2 | 2 | 2 | -100 |
|---|---|---|---|---|---|---|---|---|---|---|---|---|---|---|---|---|---|---|---|---|---|---|---|---|---|---|---|---|
| NIL | OTH | OTH | OTH | OTH | OTH | OTH | HEAD | HEAD | HEAD | HEAD | HEAD | HEAD | HEAD | HEAD | HEAD | REL | REL | TAIL | TAIL | TAIL | TAIL | TAIL | TAIL | TAIL | TAIL | TAIL | TAIL | NIL |

| to | produce |
|---|---|

Predicted Relation

**B - Relation Elicitation**

Training Example

Sleeve 41 is rigid and has {HEAD ~ a shoulder} 43 that sits against {TAIL ~ end} 321 of the outer conductor contact 37

sits against

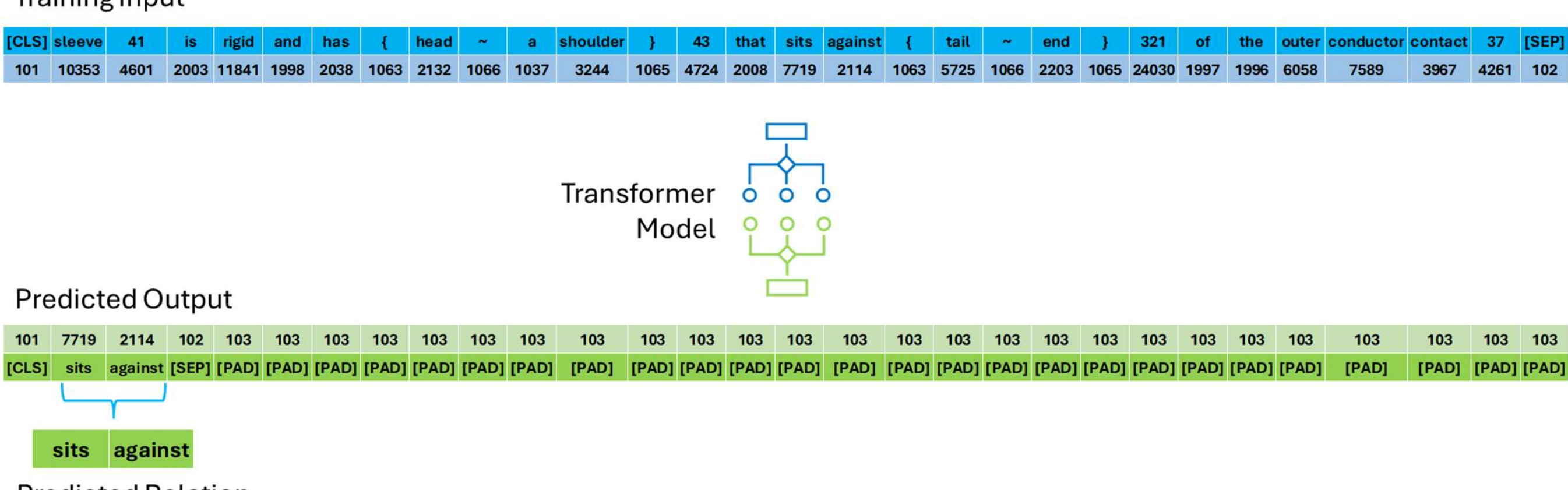

**Figure 2**: Fine-tuning language models for A) Relation Identification and B) Relation Elicitation.

In relation identification, as depicted in Figure 2A, we create training examples as tokens with labels (HEAD, REL, TAIL, and OTH) and fine-tune language models for token classification.

- Since our entities are selected from noun phrases given by spaCy transformers[36], we leverage the training module for the token tagger given by spaCy[37]. The module offers 3 choices of models as listed in Table 2 – small, large, and transformer-based (uses roBERTa-base).
- We also include the DistilBERT encoder that is commonly used for token classification tasks. Further, as listed in Table 2, we include BERT, ALBERT, and SciBERT that have been used in knowledge extraction algorithms (Ye et al., 2022; Zhong and Chen, 2021).

In relation elicitation, as depicted in Figure 2B, we form training examples as marked sentences with relation outputs and fine-tine language models for a Seq2Seq task. The models that are suitable for this task are generally heavy and fall beyond the limits of our hardware accessibility. Among feasible ones, we include BART and T5 encoders as listed in Table 2. As mentioned earlier, we make the dataset[4] and training infrastructure[5] for these models accessible.

**Table 2**: Performances of different encoders for the tasks of relation identification and elicitation. For all encoders, the dataset of 375,084 examples (187,200 positives, 187,884 negatives), we split training and testing sets by 9:1. The relation accuracy for an example is 1 if the model returns the exact relationship (if exists) or None (if not exists).

| | **Model** | **Params (M)** | **Model Loss** | **Tagger Loss** | **Rel. Accuracy** |
|---|---|---|---|---|---|
| **Relation Identification (spaCy)** | *spacy/en-core-web-sm* | - | 17.186 | 1125.017 | 0.850 |
| | *spacy/en-core-web-lg* | - | 244.156 | 3815.050 | 0.895 |
| | ***spacy/en-core-web-trf*** | **-** | **193.181** | **2285.626** | **0.954** |
| | **Encoder** | **Params (M)** | **Training Loss** | **Validation Loss** | **Rel. Accuracy** |
| **Relation Identification** | *distilbert/distilbert-base-uncased* | 67 | 0.004 | 0.004 | 0.997 |
| | *google-bert/bert-base-uncased* | 110 | 0.004 | 0.004 | 0.996 |
| | *albert/albert-base-v2* | 11.8 | 0.003 | 0.003 | 0.995 |
| | ***albert/albert-large-v2*** | **17.9** | **0.003** | **0.003** | **0.997** |
| | *allenai/scibert-scivocab-uncased* | - | 0.004 | 0.004 | 0.997 |
| **Relation Elicitation** | *facebook/bart-base* | 139 | 0.076 | 0.067 | 0.948 |
| | *google-t5/t5-small* | 60.5 | 0.144 | 0.0915 | 0.907 |
| | *google-t5/t5-base* | 222 | 0.108 | 0.0743 | 0.891 |

As observed in Table 2, relation identification that adopts a token classification task offers significantly higher relation accuracy compared to the elicitation approach. Token classification task also constraints the model to identify relation from the sentence and makes it preferable for identifying explicit relations as intended. ALBERT (large) encoder offers the best performance despite having fewer parameters. Acknowledged as a limitation by Zhong and Chen (2021, p. 5), extracting knowledge by repeatedly inputting pairs of entities is a computational challenge. For this reason, we utilise the transformer model (*en-core-web-trf*) trained using spaCy for the application purposes in this paper, while making the fine-tuned version of the ALBERT large accessible[6]. The custom-trained spaCy components could be easily included in the patent processing pipeline, and these are optimized for scalability, while also accommodating GPU acceleration. In GitHub[7], we facilitate the usage of the method by packaging a series of tasks from mining patent text to populating facts.

**3.3. Alternative Approaches**. Apart from relation identification and elicitation, we experimented link prediction approach (as depicted in Figure 3) for populating facts from a sentence. Prior knowledge extraction algorithms include link prediction as the approach to extracting relationships between entities (Sun and Grishman, 2022; Zhong and Chen, 2021). In our work, we perform link prediction between pairs of entities and relationships to construct facts. For this approach, entity and relationship tokens must be identified beforehand. Hence, we train a tagger using spaCy (as in Section 3.2) to classify tokens as "ENT", "REL" and "OTH". For 44,227 sentences split 4:1 training/testing, the resultant model exhibits tagger accuracy of 0.93 and losses being 2041.07 (transformer) and 29211.12 (tagger).

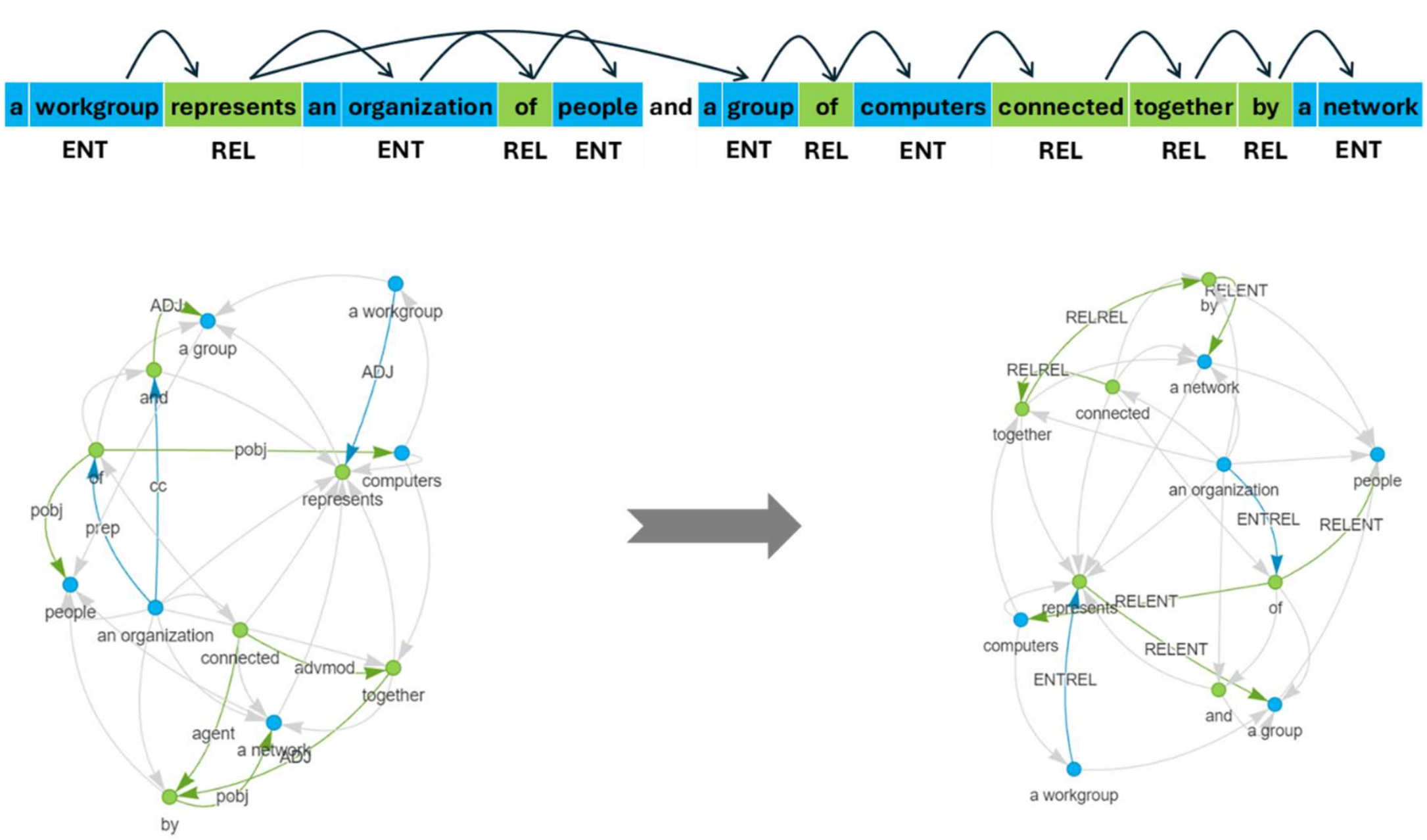

**Figure 3**: Link prediction between pairs of entities and relationships to construct facts.

As a fact includes ENT and REL tokens, the associations among the terms could be ENTREL, RELREL, and RELENT as shown in Figure 3. A pair of entity or relationship terms could be syntactically dependent and/or placed adjacent to each other. This approach examines how well these dependencies translate to associations that form facts. To predict associations between pairs of these terms, we train neural networks that include 1) Multi-Layer Perceptron[38] (MLP) and 2) various convolutional layers[39] built for Graph Neural Networks (GNNs). To incorporate node features, we fine-tune BERT for masked language modelling[40] (training loss = 0.009) and concatenate with one-hot encoding of parts of speech.

The performances of these layers summarised in Table 3 suggest that GNNs offer poor performance despite capturing the whole sentence as a graph object. While MLP offers relatively better performance, its usage on external examples reveals that prepositions like 'of' and 'at' are always associated by RELREL and with other entities, resulting in various meaningless facts. Although link prediction incorporating syntactic dependencies is adopted in the literature (Jang et al., 2021; Sun and Grishman, 2022; Zuo et al., 2022), these appear less meaningful for populating engineering design facts. Moreover, this approach is tedious in terms of training and application, unlike the relation identification that offers a significantly higher performance while also facilitating scalable application as demonstrated in the following section.

**Table 3**: Performances of various layers used in neural networks to predict links between entity and relationship pairs.

| | Entity-Relation (ENTREL) | | Relation-Relation (RELREL) | | Relation-Entity (RELENT) | |
|---|---|---|---|---|---|---|
| **No. of Edges** | 327,104 | | 169,100 | | 362,722 | |
| **No. of Graphs** | 44,098 | | 32,246 | | 44,098 | |
| **No. of Nodes** | 489,819 | | 413,800 | | 489,819 | |
| **No. of Edges** | 326,878 | | 169,100 | | 361,508 | |
| | **Loss** | **Accuracy** | **Loss** | **Accuracy** | **Loss** | **Accuracy** |
| **Muti-Layer Perceptron** | 0.284 | **0.883** | 0.101 | **0.962** | 0.18 | **0.94** |
| **Continuous Filter Convolution** | 0.520 | **0.675** | 0.183 | 0.889 | 0.456 | 0.701 |
| **Graph Convolution** | 0.691 | 0.500 | 0.666 | 0.555 | 0.692 | 0.501 |
| **PNAC (max)** | 0.536 | 0.552 | 0.192 | 0.889 | 0.481 | 0.697 |
| **PNAC (mean)** | 0.511 | **0.675** | 0.275 | 0.817 | 0.451 | 0.700 |
| **PNAC (mt3)** | 0.690 | 0.500 | 0.690 | 0.501 | 0.690 | 0.502 |
| **PNAC (var)** | 0.553 | 0.604 | 0.535 | 0.585 | 0.370 | **0.746** |
| **RGC** | 0.513 | 0.676 | 0.099 | **0.919** | 0.451 | 0.701 |
| **GraphSAGE (mean)** | 0.687 | 0.499 | 0.573 | 0.615 | 0.689 | 0.504 |
| **GraphSAGE (pool)** | 0.690 | 0.501 | 0.668 | 0.555 | 0.690 | 0.504 |

**Note**: PNAC – Principal Neighbourhood Aggregation Convolution, RGC – Relational Graph Convolution

## 4. Application

**4.1. Overview**. The motivation of this paper is to present a method to populate explicit engineering design facts from artefact descriptions so that these are utilised to guide LLMs to generate domain-specific responses during prompts. In this section, as depicted in Figure 4, we demonstrate three scenarios of GPT-4 Turbo[41] usage while retrieving generalisable and contextualised knowledge of fan systems. While Scenario 1 involves direct prompts, Scenarios 2 and 3 include references to raw text and explicit facts respectively. For this application, we recurrently apply our method to populate a knowledge base (summarised in Table 4) of over 2.9 million facts from sentences in 4,870 fan system patents under the subclass - F04D (pumps for liquids). As mentioned earlier, we make the knowledge base accessible on Hugging Face[8].

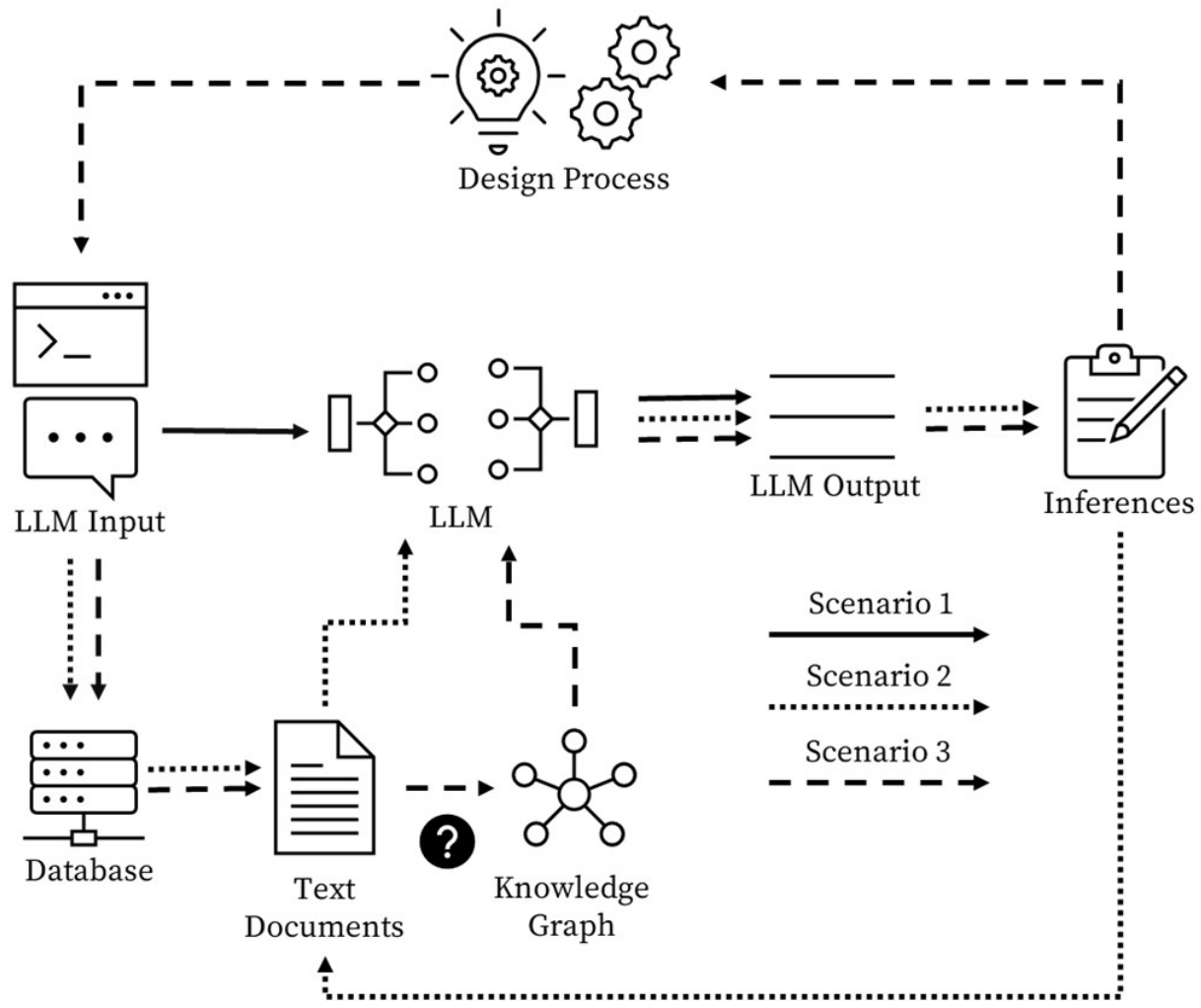

**Figure 4**: Scenarios for usage of LLMs in the design process, including RAG.

**Table 4**: Knowledge Base of Fan Systems

| | |
|---|---|
| Number of Patents | 4,870 |
| Number of Sentences | 603,184 |
| Number of Facts | 2,927,529 |
| Number of Unique Entities | 261,351 |
| Number of Unique Relationships | 115,782 |

**4.2. Generalisable Design Knowledge**. Design process occasionally requires exploring or communicating generalisable knowledge about a product domain (Luo et al., 2021). In this application, we ask the LLM to generate a technical description of the domain of fan systems in three scenarios as depicted in Figure 4.

- In Scenario 1, Table 5, when LLM is consulted directly for retrieving generalisable knowledge of fan systems, the response includes general use case, basic components, usual types, performance parameters etc. Since this response is more "Wikipedia like" and less useful for design environment, it is necessary to guide LLM with domain knowledge.
- In Scenario 2, Table 5, where textual reference is to be provided, it is not possible to input all 4,870 patents due to the token limits. We therefore provide the titles and abstracts of 10 randomly selected patents. In this scenario, the LLM locally summarises individual abstracts into short statements without offering a cohesive description of the domain.
- In Scenario 3, Table 5, upon providing 177 recurring facts as a reference, the LLM's response includes a cohesive description of a typical fan system, synthesising facts that represent generalisable attributes ("of"), hierarchy ("comprises"), behaviours ("rotatable about"), purposes ("to rotate"), and spatial arrangements ("are affixed to").

**Table 5**: Using LLMs for retrieving generalisable knowledge.

| ***Scenario 1*** |
|---|
| ***Instructions****: The output should be a single paragraph within 150 words. Provide a technical description of the domain of fan systems.*<br>***Response****: The domain of fan systems encompasses the design, engineering, implementation, and maintenance of various mechanical devices used to induce or enhance airflow within a defined environment. These systems typically involve rotary blades powered by electric motors to move air or other gases, crucial for applications like ventilation, air conditioning, drying, and exhaust purposes. Fan systems can be differentiated by factors such to include axial, centrifugal, and cross-flow, each with unique aerodynamic properties suiting specific operational requirements. Efficiency in these systems is often optimized through careful selection of fan size, shape, motor type, and control systems, which can include variable frequency drives and sensors for adaptive performance.* |
| ***Scenario 2*** |
| ***Instructions****: The context includes ten patent descriptions of a fan system. The output should be a single paragraph within 150 words. Provide a technical description of the domain of fan systems based on the following context.*<br>*Patent: 11364772. Air conditioning unit. An air conditioning unit, including: a chamber; a fan...*<br>*Patent: 11242862. Blower device. A blower device includes a rotor rotatable about a central axis...*<br>*Patent: 10327392. Battery-powered debris blower. A debris blower, having a fan frame...*<br>***Response****: The domain of fan systems, as evidenced by patents 11364772, 11242862, 10327392, 11098953, 10480520, 9217440, 10982681, 9097261, 10495114, and 10892606, demonstrates a wide array of technological advancements focused on improving efficiency, minimizing vibration, enhancing airflow, and integrating smart control systems. Innovations include specialized configurations like centrifugal and axial fans, heat exchangers, motor-driven assemblies, and structural modifications to reduce noise and vibration. These patents highlight design improvements such as thermal regulation by integrating a temperature detector, flow optimization through strategically designed blades and casing structures, and adaptations for specific applications including ceiling fans, electrical cabinet ventilation, and vehicular mounts, emphasizing the versatility and continual evolution within fan technology domains.* |
| ***Scenario 3*** |
| ***Instructions****: The context includes facts in the form – "entity :: relationship :: entity". The output should be a single paragraph within 150 words. Provide a technical description of the domain of fan systems based on the following context.*<br>*centrifugal fan :: comprises :: electric motor*<br>*fan blades :: are affixed to :: upper surface*<br>*air flow :: of :: axial fan*<br>*fan blades :: extending in :: axial direction*<br>*air flow :: generated by :: ceiling fan*<br>*fan motor :: to rotate :: fan blade*<br>*fan blades :: rotatable about :: central axis*<br>*fan blades :: extending radially from :: outer periphery...* |

***Response**: In the domain of fan systems, various types encompass several mechanical components and directional flows. Axial and centrifugal fans, powered by fan motors typically containing electric motors, encompass blades set in both radial and axial directions. The airflow is critical and may be directed longitudinally through different components and heat exchangers to facilitate air inlet and air outlet mechanisms. Central to the fan assembly, the fan blades—rotatable about a central axis—are crucial for directing air from the motor across the fan's components. Trailing edges and leading edges of fan blades play significant roles in their aerodynamic efficiency, impacting the efficiency of devices like ceiling fans or cooling fans in gas turbine engines. These fan systems are often integrated directly adjacent to heat exchangers to efficiently manage air throughput and cooling functions, optimized by motor housing that can shield vital electrical and operational components.*

To retrieve the 177 facts as represented as a knowledge graph in Figure 5, we first remove the duplicate facts within patents resulting in 2 million facts from which we identify the 30 most frequent entities (without generic terms like "method") Among 3,301 facts shared by these entities, we select the most frequent relation for each pair, resulting in 177. The availability of a knowledge base in the form of explicit facts enables us to strategically select the most representative facts of a domain and thus facilitate RAG.

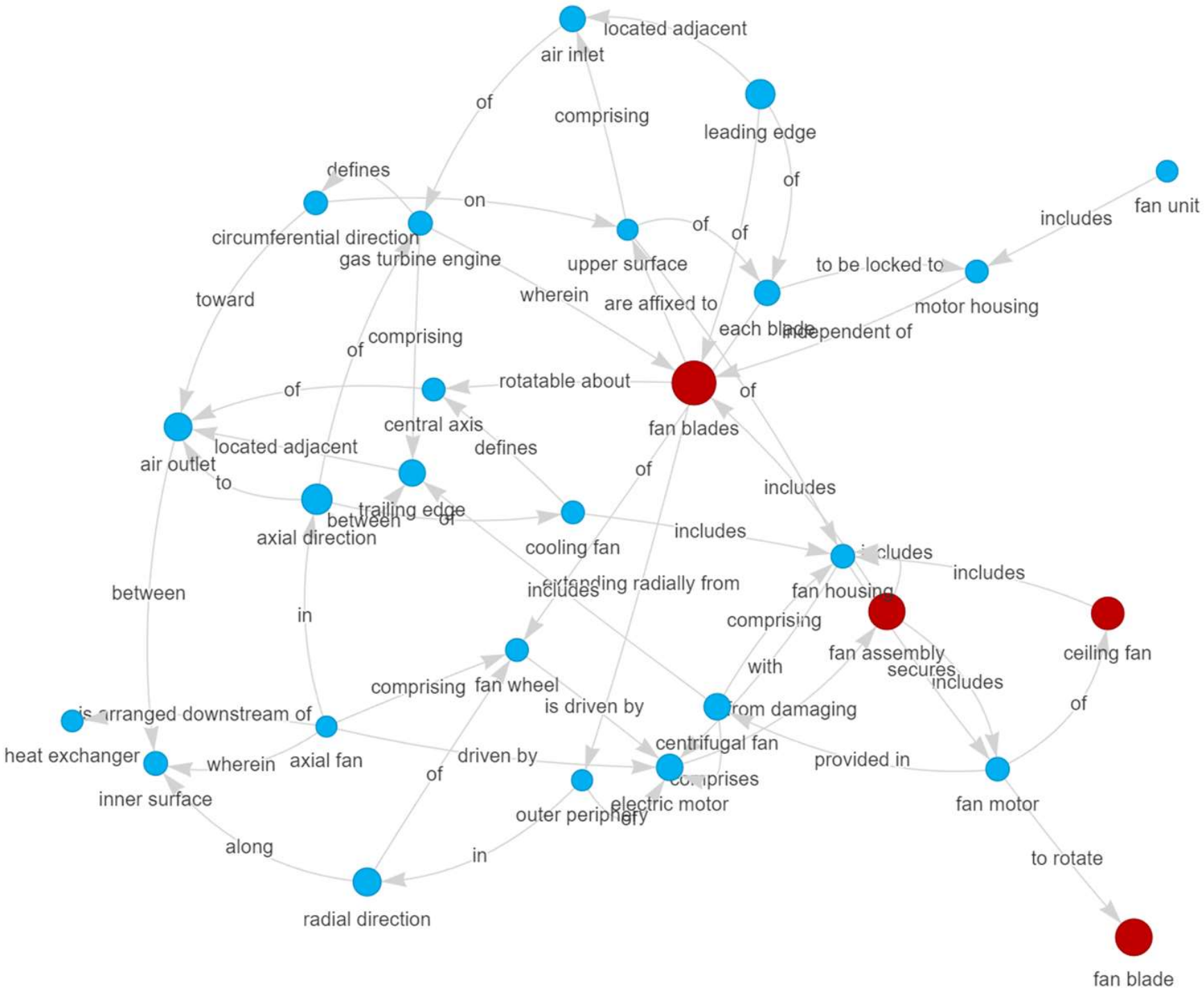

**Figure 5**: A portion of 177 recurring facts that represent generalisable design knowledge of fan systems. An interactive graph could be viewed externally[42].

**4.3. Contextualised Design Knowledge**. In a design environment, certain tasks like fault propagation require knowledge that is specific to a subsystem or an issue (Chen et al., 2022). In this application, we ask the LLM to explain the causes and solutions of "airflow noise" that is recognised as a common issue by the United States Department of Energy[43].

- In Scenario 1, Table 6, consulting LLM directly to understand the airflow noise issue results in speculative causes and solutions without reference to solutions and thus implementation details.
- In Scenario 2, Table 6, from 4,870 patents, we retrieve 559 sentences that mention "airflow" and "noise". Since the context is significantly long, the summary generated by LLM lacks traceability. In addition, the statements within the summary seem to infer noise reduction as an effect of other phenomena like airflow circulation and volume.
- In Scenario 3, Table 6, when 141 selected facts are provided as reference, LLM generates a much clearer response, indicating direct and specific solutions like serpentine flow path[44], noise reduction plate[45], sound absorbing material[46], and unique radial arrangement[47].

**Table 6**: Using LLMs for retrieving specific knowledge of issues.

| ***Scenario 1*** |
|---|
| ***Instructions**: The output should be a single paragraph within 100 words. Provide a technical description of the causes and solutions of airflow noise issue in fan systems.* <br> ***Response**: Airflow noise in fan systems primarily results from turbulence and obstruction interactions within the airflow path. When air moves past or through fan blades, disturbances such as vortices and flow separation can generate unwanted noise. Key contributors include impeller design, blade shape, and the surface roughness of components. Obstructions like grilles or poorly aligned ducts exacerbate turbulence, increasing noise levels. Solutions to mitigate this involve optimizing blade geometry using computational fluid dynamics to smooth airflow transitions, lining ducts with sound-dampening materials, and ensuring component alignment to minimize airflow disruptions. Additionally, the use of quieter, more efficient fans with variable speed controls can adapt to different airflow requirements while reducing noise output.* |
| ***Scenario 2*** |
| ***Instructions**: The context includes a list of sentences. The output should be a single paragraph within 100 words.* <br> *Provide a technical description of the causes and solutions of airflow noise issue in fan systems based on the following context.* <br> *…the vortex generators 600 may reduce noise caused in the propeller fan…* <br> *…the present invention to provide a motor fan assembly with improved airflow and noise reduction properties…* <br> *…The motor cover 114 assists in routing the cooling airflow, minimizing motor noise…* <br> ***Response**: In addressing the issue of airflow noise in fan systems, several innovative solutions are implemented to modify flow patterns and reduce turbulence, which contribute to noise. For instance, altering airflow directions at the inlet and outlet changes the noise's frequency properties and disrupts the formation of noise dipoles. Implementing flow channels and saw teeth in the design helps in managing air impacting surfaces and streamlining airflow inside the enclosure, enhancing efficiency and reducing noise. Coverage of configurations such as smaller angles in the placement of saw teeth can prevent pressure fluctuations, minimizing swirling flows and recirculating airflows; hence reducing overall noise levels. Furthermore, adjusting fan blade characteristics like blade angle, pitch, and spacing can influence airflow volume and noise levels independently, allowing for a quieter operation. These technical approaches improve fan assembly efficiency, minimize unpleasant acoustic characteristics, and maintain desired performance levels.* |
| ***Scenario 3*** |
| ***Instructions**: The context includes facts in the form - 'entity :: relationship :: entity'. The output should be a single paragraph within 100 words. Provide a technical description of the causes and solutions of airflow noise issue in fan systems based on the following context.* <br> *airflow path :: configured to attenuate :: noise* <br> *airflow :: is guided by :: noise reduction plate* <br> *unique radial arrangement :: serves to reduce :: airflow noise* <br> *outer end :: extends to reduce :: airflow noise* <br> *sound absorbing material :: for absorbing :: airflow noise…* <br> ***Response**: Airflow noise in fan systems primarily emerges from turbulence, blade tip interactions, and airflow separations. Serpentine airflow paths and noise reduction plates play pivotal roles in diminishing especially high-frequency noises by structuring the airflow to pass through carefully designed routes that attenuate sound. Integrating airflow silencing apparatuses, such as those using sound-absorbing materials or unique radial arrangements, effectively curtails noise production. Motor fan assemblies that feature improved airflow properties also reduce fan noise concurrently with cooling efficiency. Advanced methods like plasma airflow control and airflow recirculation are employed to manage and lower noise levels. Essentially, by reconfiguring airflow passages and introducing noise-reduction-focused components, these systems achieve quieter operation while maintaining desirable airflow characteristics.* |

The solutions communicated in Scenario 3 could be traced to the patents through the interactive knowledge graph[48] shown in Figure 6 that also uncovers other solutions like plasma airflow control[49]. To gather the 141 facts, we first obtain 164 facts that include the words "airflow" and "noise" in either of the entities. Using the sentence IDs of these facts, we retrieve 1,131 facts from all sentences and then converge to 141 facts that directly expand from the initial set. The availability of domain knowledge in the form of explicit facts allows us to employ these heuristics and thus enable RAG. The traceability of LLM responses to explicit facts and then to actual patents could ensure the trustworthiness of future AI systems to be deployed for engineering design applications (Díaz-Rodríguez et al., 2023).

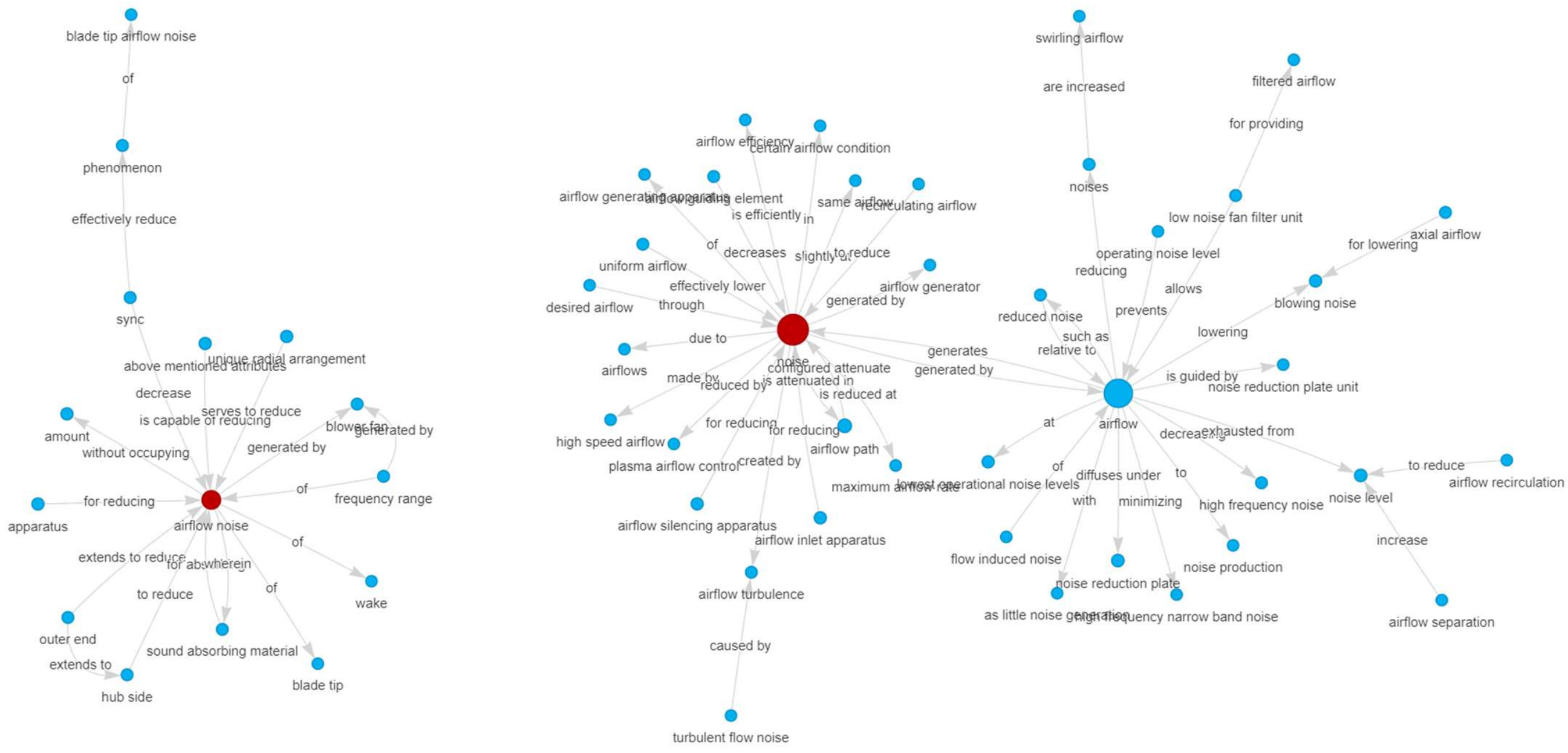

**Figure 6**: A portion of 141 facts constituting specific knowledge of the airflow noise issue fan systems. The entire knowledge graph with interaction could be viewed externally.

## 5. Conclusions

In this paper, we proposed a method to populate explicit facts of the form – head entity :: relationship :: tail entity from sentences in patented artefact descriptions. The method is primarily intended to support RAG for LLMs that are expected to be the governing platforms for future engineering design applications, especially those with knowledge-intensive tasks.

- Prior to developing this method, we created a comprehensive dataset of facts identified from 44,227 sentences that are stratified according to the CPC codes and patent document section. The dataset is our primary contribution.
- The method, which is the secondary contribution, is based on token-classification models fine-tuned on existing language models such as ALBERT, roBERTa, DistilBERT etc., that significantly outperform other approaches such as sequence-to-sequence and link prediction.
- We apply the method to 4,870 patent documents in the domain of fan-systems. We demonstrated retrieval of generalisable and contextualised design knowledge and how such knowledge can guide LLMs to generate technical and cohesive responses. Such an approach and the associated infrastructure is our tertiary contribution.

The method shall address various limitations and consider future research directions as well. We would build knowledge bases and interfaces to conduct experimental studies on how RAG aids knowledge-intensive tasks in the design process. Such experiments shall also include comparisons against well-matured retrieval approaches that include text embeddings, knowledge graph embeddings, graph-based retrieval, and other common methods such as BM25. In a design environment, we envisage that knowledge graphs as in Figures 5 and 6 are continuously navigated and explored as LLM simultaneously generates summaries of instantaneous graphs. The integration with LLM is not quantitively characterised in the current work in terms of how it benefits comprehension and creativity in the design process (Siddharth and Chakrabarti, 2018). As acknowledge earlier, repeated usage of the method for all pairs of entities in a sentence could be computationally expensive when the application is scaled for much larger set of documents. The method should therefore consider model pruning or surrogate modelling techniques for future upgrades.

---

[1] 5-cycloalkenyl 5H-chromeno[3,4-f]quinoline derivatives as selective progesterone receptor modulator compounds - https://patents.google.com/patent/US7071205

[2] United States Patent and Trademark Office (USPTO) - https://www.uspto.gov/patents/search

[3] Demo of the tagging interface - https://youtu.be/YuhLQtleZZo

[4] Dataset of engineering design facts - https://huggingface.co/datasets/siddharthl1293/engineering_design_facts

[5] Training Infrastructure for Engineering Design Knowledge Extraction - https://zenodo.org/records/12012131

[6] ALBERT Large v2 fine-tuned for token classification - https://huggingface.co/siddharthl1293/albert-albert-large-v2

[7] Engineering Design Knowledge Extraction and Usage - https://github.com/siddharthl93/engineering-design-knowledge/tree/main

[8] Knowledge of Fan Systems - https://huggingface.co/datasets/siddharthl1293/fan_systems_facts

[9] External guide for RAG - https://github.com/siddharthl93/engineering-design-knowledge/blob/main/others/RAG.md

[10] Automatic Content Extraction - https://www.ldc.upenn.edu/collaborations/past-projects/ace

[11] Workshop on Noisy and User-generated Text - https://noisy-text.github.io/2024/

[12] OntoNotes - https://catalog.ldc.upenn.edu/LDC2013T19

[13] Entity Recognizer in spaCy - https://spacy.io/api/entityrecognizer

[14] NER in TextRazor - https://www.textrazor.com/named_entity_recognition

[15] Cooperative Patent Classification Scheme - https://www.uspto.gov/web/patents/classification/cpc/html/cpc.html

[16] Method for the manufacturing of yarns from recycled carbon fibers - https://patents.google.com/patent/US20140245577A1/

[17] Patents View Data Download - https://patentsview.org/download/data-download-tables

[18] Beautiful Soup Usage - https://opensource.com/article/21/9/web-scraping-python-beautiful-soup

[19] Example Google Patents page - https://patents.google.com/patent/US6732412B2/

[20] Most frequent headings among discarded/retained - https://github.com/siddharthl93/engineering-design-knowledge/blob/main/resources/sample_headings.csv

[21] More examples of sentences and facts - https://github.com/siddharthl93/engineering-design-knowledge/blob/main/resources/selected-facts.txt

[22] Phosphite in MgX.sub.2 supported TiX.sub.3 /AlCl.sub.3 catalyst system - https://patents.google.com/patent/US4130503/

[23] Thin film magnetic recording media - https://patents.google.com/patent/US7208204/

[24] Power supply apparatus for slide door in motor vehicle - https://patents.google.com/patent/US6575760/

[25] Gas economizer - https://patents.google.com/patent/US4090466/

[26] Acid electrotinning bath - https://patents.google.com/patent/US4073701/

[27] Single crystal group III nitride articles and method of producing same by HVPE method incorporating a polycrystalline layer for yield enhancement - https://patents.google.com/patent/US8637848/

[28] Nonvolatile memory device having self refresh function - https://patents.google.com/patent/US5347486/

[29] Installation for manufacturing registration carriers - https://patents.google.com/patent/US5451155/

[30] Torque or force amplifying actuator and method for controlling actuator - https://patents.google.com/patent/US5791228/

[31] Beer keg cooling container - https://patents.google.com/patent/US4042142/

[32] Methods of treating colitis involving IL 13 and NK T cells - https://patents.google.com/patent/US8173123/

[33] Phosphite in MgX.sub.2 supported TiX.sub.3 /AlCl.sub.3 catalyst system - https://patents.google.com/patent/US4130503/

[34] Scanning files using direct file system access - https://patents.google.com/patent/US7860850/

[35] Adjustment method of optimum write power and optical write/retrieval device - https://patents.google.com/patent/US7978576B2/

[36] spaCy pipeline model built by fine-tuning transformer-based language models - https://spacy.io/models/en#en_core_web_trf

[37] spaCy training module for custom pipeline components - https://spacy.io/usage/training#quickstart

[38] MLP Classifier - https://scikit-learn.org/stable/modules/generated/sklearn.neural_network.MLPClassifier.html

[39] DGL Convolutional Layers - https://docs.dgl.ai/api/python/nn-pytorch.html

[40] BERTForMaskedLM - https://huggingface.co/docs/transformers/en/model_doc/bert#transformers.FlaxBertForMaskedLM

[41] GPT-4 Turbo - https://platform.openai.com/docs/models/gpt-4-turbo-and-gpt-4

[42] Generalisable Design Knowledge of Fan Systems - https://fansystems.vercel.app/

[43] Improving Fan System Performance - https://www.nrel.gov/docs/fy03osti/29166.pdf

[44] Air blower assembly for vacuum cleaner - https://patents.google.com/patent/US4735555A/

[45] Low-noise fan-filter unit - https://patents.google.com/patent/US6217281B1/

[46] Blower - https://patents.google.com/patent/US8408865B2/

[47] Fan inlet flow controller - https://patents.google.com/patent/US5979595A/

[48] Knowledge of airflow noise in fan systems - https://fansystems.vercel.app/airflownoise

[49] Voltage application device, rotation apparatus and voltage application method - https://patents.google.com/patent/US8937799B2/